\documentclass[conference]{IEEEtran}
\IEEEoverridecommandlockouts
\usepackage{amsmath,amssymb,amsfonts}
\usepackage{algorithmic}
\usepackage{graphicx}
\usepackage{textcomp}
\usepackage{xcolor}
\usepackage{cite}

\usepackage{hyperref}
\usepackage{comment}
\begin{document}

\title{Zero-Shot Grasping from Local Surface Geometry\\
\thanks{We would like to acknowledge Natural Sciences and Engineering Research Council of Canada (NSERC) for their financial support of this work.}
}

\author{
Jayalekshmi Jayakumar\hspace{1 cm}
Eric Haden\hspace{1 cm}
Matthew Pan\hspace{1 cm}
Michael Greenspan \\
\textit{Department of Electrical and Computer Engineering} \\
\textit{Ingenuity Labs, Queen's University} \\
Kingston, Ontario, Canada \\
\{jayalekshmi.j, eric.haden, matthew.pan, michael.greenspan\}@queensu.ca
}

\maketitle
\begin{abstract}
Robotic grasping of novel objects remains challenging, especially in unstructured environments without object models or prior knowledge. Many learning-based methods rely on global representations or large datasets, limiting generalization. We present RCVGrasp, a zero-shot grasp-prediction framework that operates solely on local surface geometry. A parallel-jaw grasp is represented as a pair of 3D surface patches corresponding to gripper contacts, and grasp feasibility is formulated as a binary classification over patch pairs. Our network is trained entirely on procedurally generated synthetic data using geometric constraints, without object meshes, CAD models, or RGB information. The model learns to identify graspable contact configurations directly from point cloud geometry. Experiments demonstrate high accuracy on synthetic data, effective transfer to standard object benchmarks, and reliable real-world grasping of unseen objects. These results highlight the power of local geometry-driven representations for zero-shot, transferable grasp prediction. The dataset and implementation are publicly available at https://github.com/JJayaL/RCVGrasp.
\end{abstract}


\section{Introduction}

Robotic grasp prediction remains a fundamental challenge in manipulation, particularly in unstructured environments with previously unseen objects. Although learning-based methods have achieved strong performance, many rely on object-specific information such as object models, global scene context, or dense grasp supervision \cite{yan2018learning,sundermeyer2021contact,fang2020graspnet,morrison2018closing}, which can limit generalization beyond the training distribution \cite{newbury2023deep}. These dependencies are often reinforced by simulation-heavy training pipelines and object-aware formulations, which frequently require sim-to-real adaptation to bridge the out-of-distribution gap \cite{holst2022generation,liu2023digital,chisari2024centergrasp, chen2022towards}. While recent work has shown that reconstructing scene geometry can improve collision-aware grasping in clutter via contact- and reconstruction-based refinement \cite{iwase2025zerograsp}, achieving reliable zero-shot grasping without prior object knowledge remains an open problem. In contrast, analytic and geometry-driven methods emphasize local surface properties at contact regions—such as curvature, antipodal alignment, and force closure \cite{kopicki2016one,murali2018learning}—offering object-independent and interpretable formulations, but often suffering from brittleness under sensor noise, partial observations, and contact-induced object motion \cite{zhou2017probabilistic}.

\begin{figure}[tbp]
    \centering
    \includegraphics[width=0.5\textwidth]{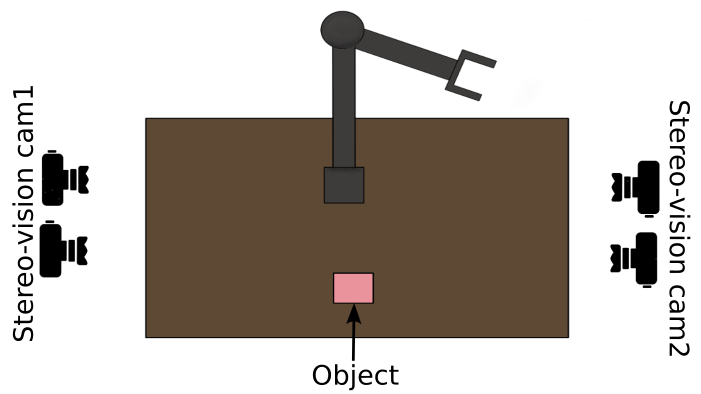}
    \caption{System setup}
    \label{fig:system_setup}
\end{figure}

\begin{figure}[tbp]
    \centering
    \includegraphics[width=0.5\textwidth]{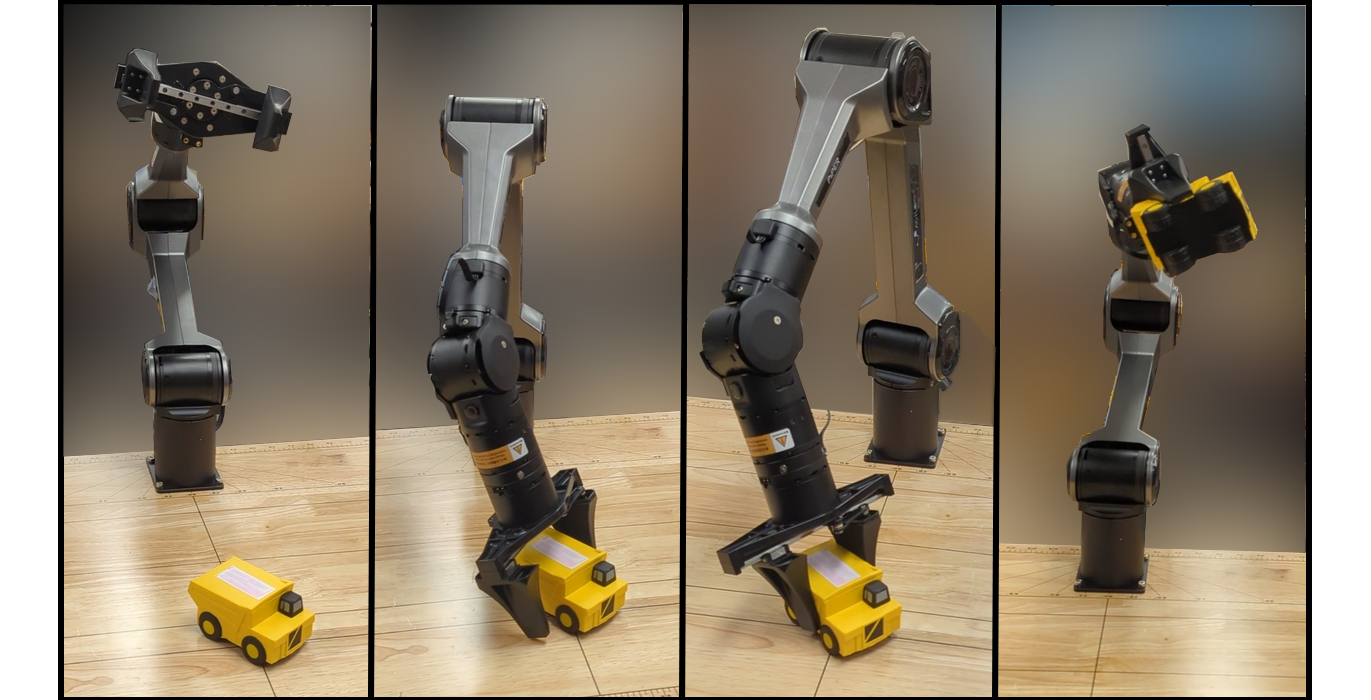}
    \caption{Real-world zero-shot grasp using local surface geometry from stereo point clouds.}
    \label{fig:zero_shot_grasp}
\end{figure}

Motivated by these limitations, we revisit grasp prediction from a purely local geometric perspective. Rather than predicting full grasp poses or working with entire object shapes, we focus on the feasibility of grasp contacts themselves, similar to contact-anchored representations \cite{sundermeyer2021contact}, but without requiring grasp pose regression or dense grasp annotations. Specifically, we model a parallel-jaw grasp as a pair of local surface patches corresponding to the two gripper contact regions. Graspability is determined by evaluating whether these patches satisfy the geometric constraints required for stable parallel-jaw grasping. This formulation separates grasp prediction from object identity, enabling reasoning based solely on local surface geometry. Figure~\ref{fig:zero_shot_grasp} shows an example of a successful real-world grasp produced by our method.

Building on this perspective, we introduce RCVGrasp, a zero-shot grasp prediction framework that reasons over local 3D surface geometry rather than object-level information. 
Our sensing system is based on two antipodal stereovision cameras,
as illustrated in Figure~\ref{fig:system_setup}. 
This multi-view stereo setup targets industrial bin-picking tasks and complements common single-view pipelines.
The method operates directly on local point cloud regions and is trained using purely synthetically generated data derived from geometric constraints, without relying on object meshes, CAD models, or visual appearance cues. Prior work has shown that procedurally generated data grounded in geometric consistency can support effective sim-to-real transfer \cite{tobin2017domain,sundermeyer2021contact,mahler2017dex,holst2022generation,lin2020using}. By exposing the model to systematically constructed local configurations that reflect both graspable and non-graspable conditions, the framework learns discriminative geometric cues while remaining independent of object shape and identity.

At test time, the learned model is applied zero-shot to previously unseen objects or scenes to evaluate grasp feasibility based solely on local surface structure, enabling dense, object-agnostic grasp prediction. Experimental results demonstrate strong generalization from synthetic training data to real-world geometries, supporting robust zero-shot grasping and aligning with recent efforts toward transferable, geometry-informed grasp representations \cite{li2024shapegrasp}.

The main contributions of this work are:
\begin{itemize}
    \item Formulation of a local patch-pair grasp representation for parallel-jaw contacts, independent of object identity and global shape.
    \item A procedurally generated synthetic dataset of graspable and non-graspable patch pairs built from explicit geometric constraints without object models or manual annotations.
    \item A zero-shot grasp prediction framework that generalizes from patch-pair classification to object-level grasp detection on unseen geometries.
\end{itemize}

Our synthetic dataset generation code and dataset are publicly available at:
\url{https://github.com/JJayaL/RCVGrasp}.

\section{Related Work}
Prior work on robotic grasping has explored a wide range of grasp representations that trade off expressiveness, computational efficiency, and robustness in unstructured environments. Grasp notations have evolved from analytical formulations based on rigid-body mechanics and force-closure analysis to data-driven~\cite{nguyen1988constructing,ferrari1992planning}, vision-based representations, largely enabled by convolutional neural networks (CNNs) operating on RGB and RGB-D inputs~\cite{lenz2015deep,ten2017grasp}. Modern systems increasingly combine geometric reasoning with deep learning~\cite{liang2019pointnetgpd} for robust grasping in applications such as bin picking, warehouse automation, and service robotics. 
Synthetic-data toolchains and simulation-trained policies have been proposed to improve data efficiency and sim-to-real transfer~\cite{holst2022generation,mahler2017dex,lin2020using,chisari2024centergrasp,chen2022towards}, though they typically require careful sim-to-real transfer. Current research emphasizes compact, generalizable representations that avoid reliance on full object models while remaining effective in cluttered scenes~\cite{xu2022gknet}, alongside standardized benchmarks for reproducible evaluation~\cite{calli2015benchmarking,fang2020graspnet}. Recent work has also explored foundation models for zero-shot grasp reasoning, combining geometric abstractions with language-based commonsense inference~\cite{li2024shapegrasp}.



Large-scale datasets have been central to standardizing grasp representations by defining how grasp candidates are parameterized in image or 3D space. Early benchmarks such as the Cornell Grasping Dataset~\cite{jiang2011efficient} represented grasps as 2D rectangles on RGB-D images, enabling CNN-based binary grasp-feasibility prediction but lacking explicit grasp-quality modelling. The Jacquard dataset~\cite{depierre2018jacquard} introduced a structured image-space parameterization \( g = (x, y, h, w, \theta) \) and large-scale supervision, substantially advancing learning-based grasp detection. Subsequent works incorporated continuous grasp quality estimation~\cite{yu2022se} and extended representations to 3D and 6-DoF formulations~\cite{zhao2021regnet,chen2024efficient,fang2020graspnet,mousavian20196}, as well as sparse keypoint-based grasp primitives to reduce ambiguity and improve efficiency~\cite{xu2022gknet}.

\subsection{Grasp Techniques}
Early learning-based grasp detectors relied on global object features and anchor-based representations in image or point cloud space, enabling efficient proposal generation but often requiring dense supervision and large annotated datasets~\cite{lenz2015deep,ten2017grasp,cheng2023anchor}. Classical geometry-driven methods, in contrast, identify grasp positions based on an object's surface structure and often require an explicit object model and 6-DoF pose. Some approaches discretize object surfaces into hierarchical patches~\cite{jia2025patch} using differential geometry, creating patch trees with Gauss maps and PCA to capture surface normal variation and torque constraints. Grasp synthesis searches and prunes patch combinations to achieve force-closure on curved objects, but many methods rely on brute-force search and idealized assumptions.

More recent learning-based approaches extract local 3D patches from RGB-D images, normalize them to a region-specific feature space, and process them with neural networks to predict grasp poses~\cite{zhao2021regnet}. Fully convolutional grasp-map predictors regress dense grasp-quality, orientation, and gripper-width maps directly from depth or RGB-D inputs~\cite{yu2022se,morrison2018closing,tian2022rgbd}, while RGB-D dense-fusion generative methods further improve pixel-wise grasp prediction by explicitly integrating color and depth features throughout the network~\cite{tian2022rgbd}. Rather than reasoning over full object shapes, contact-centric representations focus on the feasibility of local grasp contacts~\cite{sundermeyer2021contact,kopicki2016one}, which can be applied in zero-shot parallel-jaw settings. Active perception approaches combine dense grasp prediction with viewpoint selection to handle occlusions and depth uncertainty, progressively refining grasp confidence in cluttered environments~\cite{wu2023information}.

\subsection{Grasp Evaluation and Benchmarking Metrics}

To enable fair comparison across these diverse grasping approaches, standardized evaluation metrics have been developed.
Cornell-style grasp detection methods \cite{jiang2011efficient} are commonly evaluated using an oriented rectangle matching criterion. A predicted grasp $g_p$ is considered correct if the angular difference from the corresponding ground-truth grasp $g_t$ is less than $30^\circ$ and the Jaccard index (intersection over union, IoU) \cite{depierre2018jacquard} between the two grasp rectangles satisfies:
\[
J(g_p, g_t) = \frac{|g_p \cap g_t|}{|g_p \cup g_t|} > 0.25 .
\]
This protocol enables standardized comparison across methods that output oriented grasp rectangles and is typically reported alongside inference speed measured in frames per second (fps).

While rectangle-based metrics such as IoU and orientation error are effective for evaluating image-space grasp detection, they do not fully capture manipulation performance in real-world settings. To address this limitation, object-set–based benchmarks such as the YCB Object and Model Set \cite{calli2015benchmarking} emphasize standardized physical objects and task-level evaluation protocols, using metrics including grasp success rate, task completion time, and robustness across repeated trials. These benchmarks complement grasp-level metrics by enabling reproducible, system-level comparison across perception, planning, control, and hardware configurations.

\subsection{Our Positioning}
Prior foundation-model and policy-driven grasping methods focus on broad generalization, often leveraging language reasoning, global context, or manipulation policies~\cite{li2024shapegrasp,yang2025transferring}, and typically require large datasets or object-specific priors to achieve zero-shot performance.

In contrast, our method targets zero-shot grasp feasibility using purely local surface geometry. By modelling parallel-jaw grasps as pairs of local surface patches, we evaluate graspability independently of object identity, pose, or global shape. This contact-centric formulation enables dense evaluation of grasp candidates directly from point clouds and allows training entirely on synthetic, procedurally generated data. Our approach, therefore, combines object-independent geometric reasoning with zero-shot generalization, providing an interpretable and robust alternative to task- or policy-driven methods.

\section{Approach}
\begin{figure}[t]
    \centering
    \includegraphics[width=0.5\textwidth]{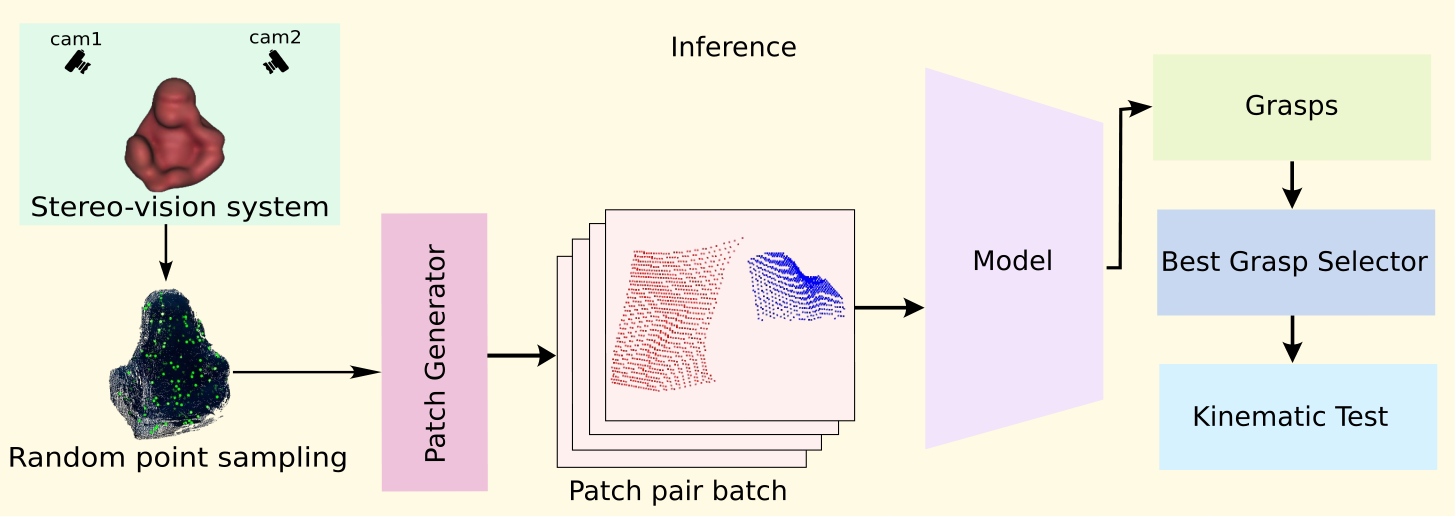}
    \caption{System architecture}
    \label{fig:system_architecture}
\end{figure}
We propose RCVGrasp, a zero-shot, local-geometry-based grasp-prediction framework that operates directly on 3D surface information. Rather than relying on object identity, models, or global scene context, our method evaluates grasp feasibility using pairs of local surface patches corresponding to the contact regions of a parallel-jaw gripper. This enables a system that does not require a prior object model or pose estimate, yet generalizes to grasp previously unseen objects. 

The system architecture is shown in Fig. \ref{fig:system_architecture}. During training, a PointNet++~\cite{qi2017pointnet++} backbone is fed pairs of synthetic point clouds representing possible surface patches of an object to be grasped. These synthetic patch pairs are randomly generated within a range of geometric constraints, and are assigned a ground truth label of either
\textit{positive} 
or \textit{negative}
indicating whether or not they are respectively
\textit{graspable}
or 
\textit{non-graspable}
by a particular parallel gripper.
During training, the network learns to classify these patch pairs,
and at inference, 
patch pairs extracted from a sensed object within a scene are then 
similarly classified as either \textit{graspable} or \textit{non-graspable}.


\subsection{Grasp Constraints and Representation}
\begin{figure}[t]
    \centering
    \includegraphics[width=0.5\textwidth]{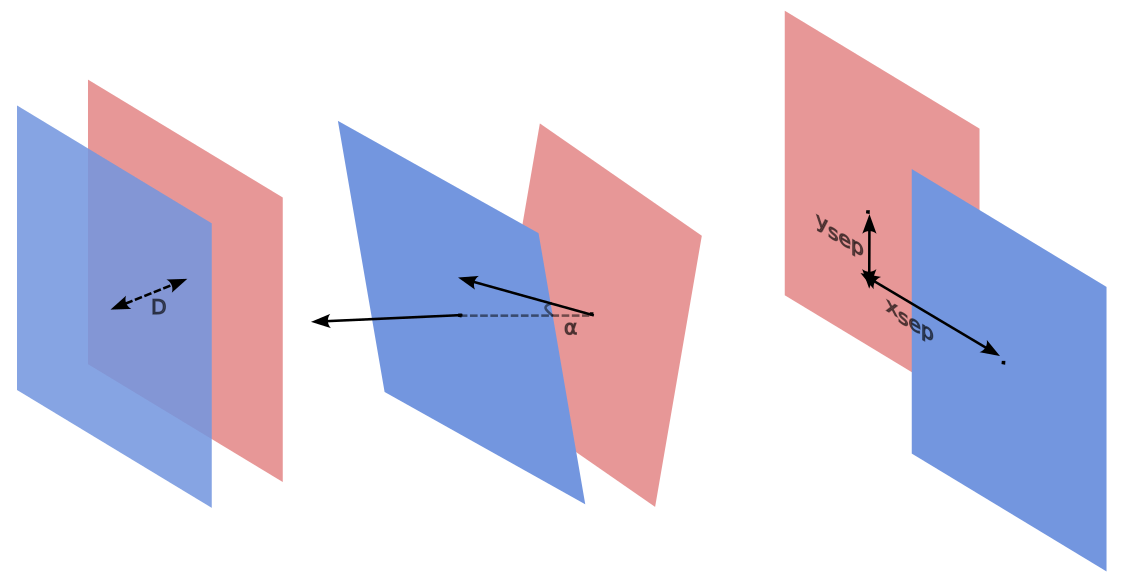}
    \caption{Distance, angle and alignment grasp criteria}
    \label{fig:criteria}
\end{figure}

Grasp constraints are determined based on the geometric characteristics of a particular parallel jaw gripper, or family of grippers.
The particular criteria (as shown in Figure~\ref{fig:criteria}) considered are:
\begin{itemize}
    \item \textbf{Distance}: The Euclidean distance between patch centers lies within a fixed range ${\mathcal{D}}=[D_{min},D_{max}]$, which corresponds to the minimum and maximum jaw separation; 
    \item \textbf{Angle}: The angle between the surface normals of the two patches is within a range $\alpha$, such that the patches are substantially antipodal (i.e., nearly parallel), mirroring the parallel configuration of the two parallel jaw gripper pads; 
    \item \textbf{Alignment}: The centers of the two patches are approximately aligned to reflect feasible parallel-jaw gripper contacts. Specifically, the two vectors, antiparallel to each gripper pad and intersecting at their centers, will pass each other within some Euclidean distance $A$.
\end{itemize}

Grasps are represented as pairs of local surface patches corresponding to the two contact regions (\textit{pads}) of a parallel-jaw gripper. Each contact region is modelled as a local 3D surface patch captured in the camera coordinate frame and represented as a 32×32×3 tensor, where the three channels store the XYZ coordinates. This representation preserves the local geometric structure of the object surface while remaining independent of object identity or appearance.
The 32×32 spatial resolution defines the extent of the local neighbourhood around a selected surface point, with the center of each patch corresponding to the predicted gripper contact location. Both patches in a grasp pair are expressed in the same camera coordinate frame to ensure geometric consistency between the two contacts.
Each grasp candidate is further accompanied by
a scalar grasp quality metric, where lower values indicate higher grasp feasibility. By relying solely on local surface geometry at the contact regions, this representation enables category-independent grasp evaluation without requiring global object shape information.

\subsection{Synthetic Dataset Construction}

The synthetic dataset is generated based on the above-defined grasp constraints. 
Pairs of 3D surface patches are generated to either satisfy or violate the gripper’s graspability criteria, resulting in graspable and non-graspable samples.
To generate synthetic training patch pairs, we procedurally sample two opposing 
point surface patches that mimic feasible contacts for a parallel-jaw gripper. Each sample consists of two point clouds, each containing $32\times 32=1024$ points. 

\textbf{Graspable patch generation.}
To generate a graspable patch pair, two points are first randomly selected.
The first point
$p_1 = (x_1,y_1,z_1)$
defines the center of the first patch, and is randomly generated to lie within a fixed volume $(x_1,y_1,z_1)\in[-4,4]\times[-3,6]\times[80,10]\ \mathrm{cm}$.
The second point 
$p_2 = (x_2,y_2,z_2) 
= (x_1\!+\!x_{\text{sep}},
y_1\!+\!y_{\text{sep}},
z_1\!+\!z_{\text{sep}})$
defines the center of the second patch,
and
is obtained by adding a random 3D offset
$||(x_{\text{sep}}, y_{\text{sep}},z_{\text{sep}})||\in{\mathcal{D}}$
to $p_1$.

\begin{figure*}[t]
    \centering
    \begin{minipage}{0.19\textwidth}
        \centering
        
        \includegraphics[width=\linewidth]{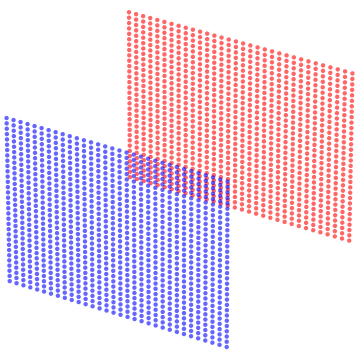}
        \textbf{(a)}\\[-0.3ex]
    \end{minipage}\hfill
    \begin{minipage}{0.19\textwidth}
        \centering
        \includegraphics[width=\linewidth]{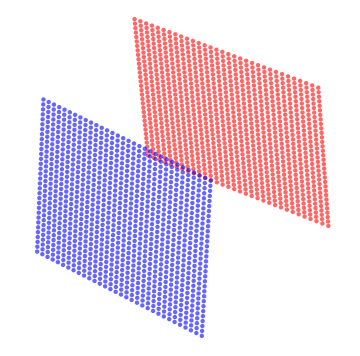}
        \textbf{(b)}\\[-0.3ex]
    \end{minipage}\hfill
    \begin{minipage}{0.19\textwidth}
        \centering
        \includegraphics[width=\linewidth]{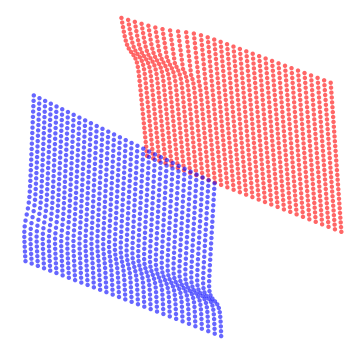}
        \textbf{(c)}\\[-0.3ex]
    \end{minipage}\hfill
    \begin{minipage}{0.19\textwidth}
        \centering
        \includegraphics[width=\linewidth]{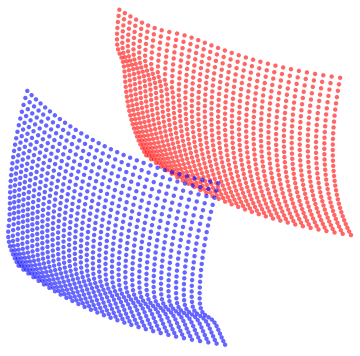}
        \textbf{(d)}\\[-0.3ex]
    \end{minipage}\hfill
    \begin{minipage}{0.19\textwidth}
        \centering
        \includegraphics[width=\linewidth]{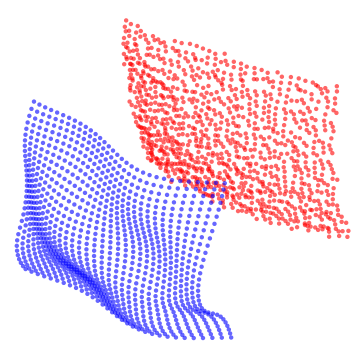}
        \textbf{(e)}\\[-0.3ex]
    \end{minipage}

    \caption{Different types of augmentations applied to generate graspable patches.
    (a) no augmentation,
    (b) slight tilt,
    (c) added bend,
    (d) added curvature,
    (e) wave-like texture on the blue patch and noise added to the red patch.}
    \label{fig:good_patch}
\end{figure*}

\begin{figure*}[t]
    \centering
    \begin{minipage}{0.19\textwidth}
        \centering
        \includegraphics[width=\linewidth]{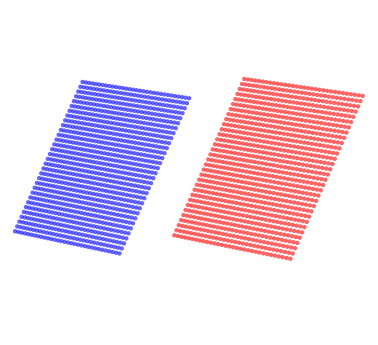}
        \textbf{(a)}\\[-0.3ex]
    \end{minipage}\hfill
    \begin{minipage}{0.19\textwidth}
        \centering
        \includegraphics[width=\linewidth]{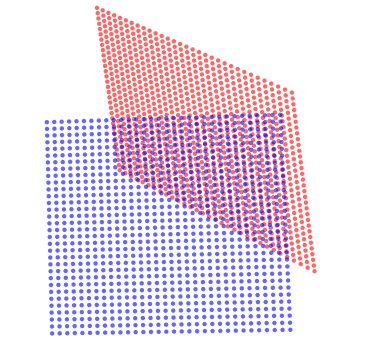}
        \textbf{(b)}\\[-0.3ex]
    \end{minipage}\hfill
    \begin{minipage}{0.19\textwidth}
        \centering
        \includegraphics[width=\linewidth]{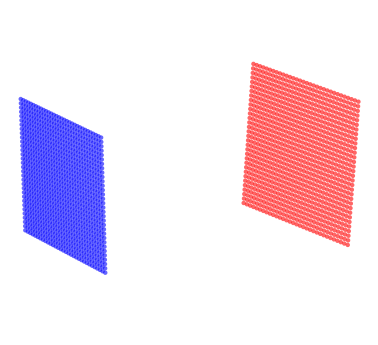}
        \textbf{(c)}\\[-0.3ex]
    \end{minipage}\hfill
    \begin{minipage}{0.19\textwidth}
        \centering
        \includegraphics[width=\linewidth]{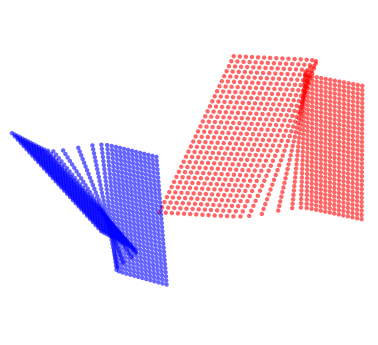}
        \textbf{(d)}\\[-0.3ex]
    \end{minipage}\hfill
    \begin{minipage}{0.19\textwidth}
        \centering
        \includegraphics[width=\linewidth]{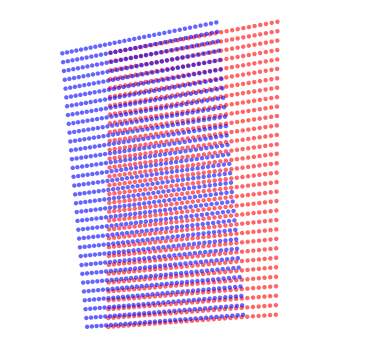}
        \textbf{(e)}\\[-0.3ex]
    \end{minipage}

    \caption{Examples of non-graspable patch pairs used in the synthetic test dataset:
    (a) co-planar patches,
    (b) patches at a large relative angle,
    (c) poor patch alignment,
    (d) excessive bending,
    (e) insufficient gap between patches.}
    \label{fig:bad_patch}
\end{figure*}

\phantomsection
\label{Graspable_patch}
This yields two parallel patches with a constrained relative displacement. Each patch is a uniformly sampled $32\times 32$ planar point grid. The local geometry of each patch is then augmented (as shown in Figure~\ref{fig:good_patch}) as follows:

\begin{itemize}
\item 
A small random tilt ($\pm 8$ degrees) along one or both axes. These augmentations are bounded so that the two surface normals still remain nearly parallel. The angle between the two patches is calculated, and if it exceeds 8 degrees, it is reduced.
\item 
Optional surface texture (e.g., noise, waves, or none).  
\item 
Optional spherical-like curvature (convex/concave and variants). 
\item 
A mild ``bend'' augmentation is applied over part of a patch, introducing spatially varying local tilt beyond the global constraint. 
\item 
Both patches are jointly rotated about their common midpoint by a rigid transform applied to the concatenated pair, preserving the relative configuration while increasing the diversity of the patches from a particular viewpoint.
\end{itemize}

After augmentation, the common centroid is between the two patches translated to the origin, ensuring each pair is centred consistently across the dataset. Overall, this procedure produces synthetic pairs that satisfy the core graspability heuristics for a parallel gripper. The output for each sample is stored as two XYZ point clouds 
providing clean, geometry-consistent positive examples suitable for training and evaluation.


\textbf{Non-graspable patch generation.}
To generate non-graspable training patch pairs, we follow the same procedural framework as for graspable samples but systematically violate one or more core graspability criteria.
As in the graspable case, two patch centers are first sampled. The first center $p_1$
is randomly generated within a fixed volume
$(x_1, y_1, z_1) \in [-4, 4] \times [-4, 4] \times [30, 65]\,\mathrm{cm}$.
The second center $p_2$
is obtained by adding a relative offset to $p_1$, which is chosen so as to intentionally violate one or more of the above-described graspability constraints. 
In this way, graspability violations (as shown in Figure~\ref{fig:bad_patch}) can result from combinations of modifications to an individual patch geometry, or to the relative positions of two patches, as follows:



\begin{itemize}
    \item \textbf{Same-plane violation:} The separation along the z-axis is constrained to $0$--$10\,\mathrm{mm}$, placing the
    two patches effectively on the same plane. While the patches may be laterally offset along the x- and y-axes (typically $20$--$50\,\mathrm{mm}$),
    the insufficient patch separation violates the minimum distance criterion $D_{min}$ required for parallel-jaw engagement.

    \item \textbf{Large-angle violation:} The angle between the surface normals of the two patches is forced to exceed
    $30^\circ$, and may extend up to $90^\circ$ (perpendicular surfaces). The geometric angle between the plane normals is explicitly computed and enforced to lie outside the allowable range $\alpha$, violating the near-parallel requirement.

    \item \textbf{Excessive alignment violation:} The horizontal
    ($XY$-plane) offset $(x_{\text{sep}}, y_{\text{sep}})$
    between patch centers is sampled from
    $35$--$90\,\mathrm{mm}$, exceeding the alignment tolerance of a parallel-jaw gripper. Although the patches may remain parallel, their relative displacement violates the alignment criterion.

    \item \textbf{Too-close separation violation:} The patches are separated by only $1$--$10\,\mathrm{mm}$ in all dimensions, producing configurations that are neither coincident nor sufficiently separated for grasping, again violating the distance criterion.

    \item \textbf{Patch geometry violation:} 
One or both patches may be bent over more than
    $70\%$ of their surface area, with local tilt variations of
    $15$--$40^\circ$, yielding highly non-graspable geometries.
    \item \textbf{Multi-violation configurations:} Multiple violations are random combinations of the above grasp violations, yielding highly non-graspable geometries that violate distance, angle, and alignment constraints simultaneously.
\end{itemize}

For most of the non-graspable samples, the same geometric augmentations used for graspable pairs (as described earlier in Section \ref{Graspable_patch}) are applied, including optional surface textures (e.g., noise or waves), spherical-like curvature, and bend augmentations. A few patches with extreme texture and curvature are also generated to increase diversity. Finally, both patches are jointly rotated about their common midpoint using a rigid transform applied to the concatenated pair, preserving their relative configuration while increasing viewpoint diversity. The combined centroid is then translated to the origin to ensure consistency across the dataset.



A synthetic dataset was generated using the procedure described above. The dataset comprised 6750
graspable and 6,750 non-grasable patch pairs, partitioned into 10,000 training, 2,500 validation and 1,000 test pairs.
The dataset was synthesized completely
from first principles and made no reference to any specific object geometry used in subsequent testing.

Grasp candidates are generated by pairing points within the same camera view and across camera views, producing both single-view and cross-view patch pairs. These points serve as a reference for the patch's center. All patches are transformed into a common camera coordinate frame prior to pairing. Each patch pair is evaluated by the trained network to predict a grasp quality score using a binary cross-entropy loss, with lower scores indicating stronger grasp candidates.

This exhaustive pairing strategy enables dense evaluation of grasp quality across object surfaces and viewpoints, supporting robust grasp prediction in cluttered and partially observed scenes.

\subsection{Training}
The network outputs a scalar graspability score, which is thresholded to produce a binary classification indicating whether the input patch pair is graspable or non-graspable for the parallel-jaw gripper modelled through the geometry constraints.
We evaluate Binary Cross-Entropy (BCE) loss during training.
We train the network using the Adam optimizer with an initial learning rate of 1e-3. Training was performed for 250 epochs with a batch size of 32, and gradient clipping with a maximum norm of 1.0 is applied to improve stability. A StepLR schedule was used to decay the learning rate by a factor of 0.7 at fixed intervals derived from a decay step of 50,000 batches, with updates applied once per epoch.
\begin{figure*}[!t]
    \centering

    \begin{minipage}{0.18\textwidth}
        \centering
        \textbf{(a)}\\[-0.3ex]
        \includegraphics[height=3cm, keepaspectratio]{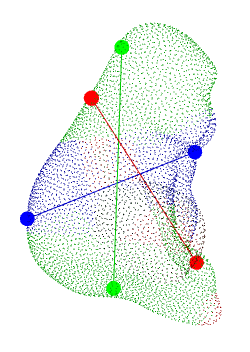}
    \end{minipage}
    \hfill
    \begin{minipage}{0.18\textwidth}
        \centering
        \textbf{(b)}\\[-0.3ex]
        \includegraphics[height=3cm, keepaspectratio]{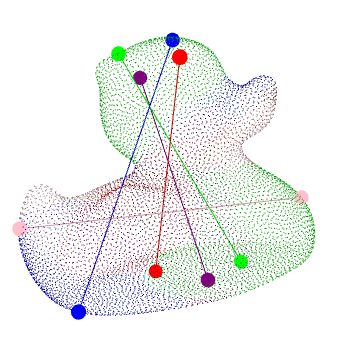}
    \end{minipage}
    \hfill
    \begin{minipage}{0.18\textwidth}
        \centering
        \textbf{(c)}\\[-0.3ex]
        \includegraphics[height=3cm, keepaspectratio]{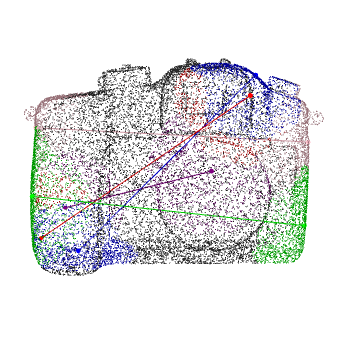}
    \end{minipage}
    \hfill
    \begin{minipage}{0.18\textwidth}
        \centering
        \textbf{(d)}\\[-0.3ex]
        \includegraphics[height=3cm, keepaspectratio]{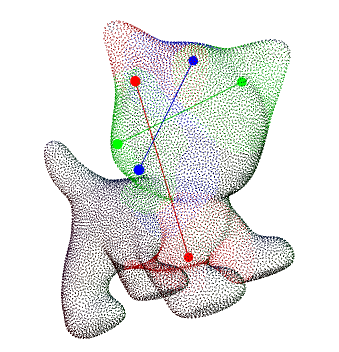}
    \end{minipage}
    \hfill
    \begin{minipage}{0.18\textwidth}
        \centering
        \textbf{(e)}\\[-0.3ex]
        \includegraphics[height=3cm, keepaspectratio]{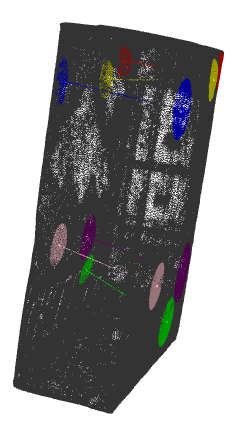}
    \end{minipage}

    \vspace{1ex}

    \begin{minipage}{0.18\textwidth}
        \centering
        \textbf{(f)}\\[-0.3ex]
        \includegraphics[height=3cm, keepaspectratio]{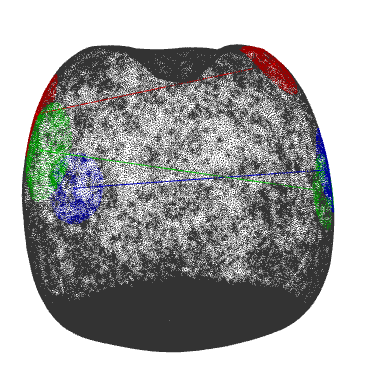}
    \end{minipage}
    \hfill
    \begin{minipage}{0.18\textwidth}
        \centering
        \textbf{(g)}\\[-0.3ex]
        \includegraphics[height=3cm, keepaspectratio]{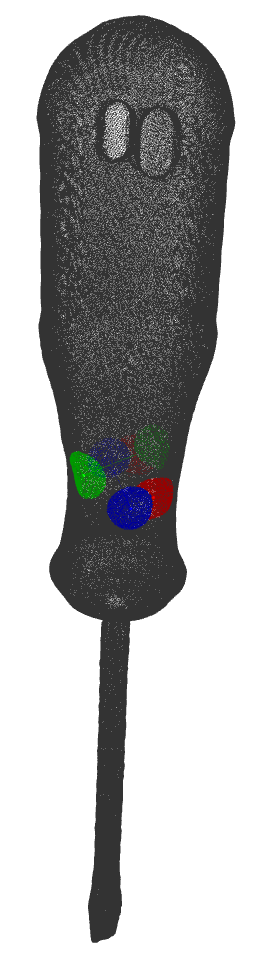}
    \end{minipage}
    \hfill
    \begin{minipage}{0.18\textwidth}
        \centering
        \textbf{(h)}\\[-0.3ex]
        \includegraphics[height=3cm, keepaspectratio]{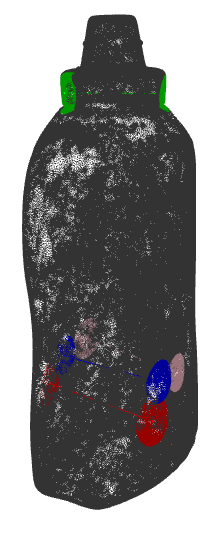}
    \end{minipage}
    \hfill
    \begin{minipage}{0.18\textwidth}
        \centering
        \textbf{(i)}\\[-0.3ex]
        \includegraphics[height=3cm, keepaspectratio]{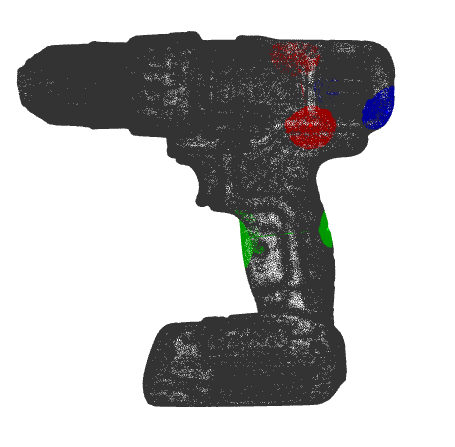}
    \end{minipage}
    \hfill
    \begin{minipage}{0.18\textwidth}
        \centering
        \textbf{(j)}\\[-0.3ex]
        \includegraphics[height=3cm, keepaspectratio]{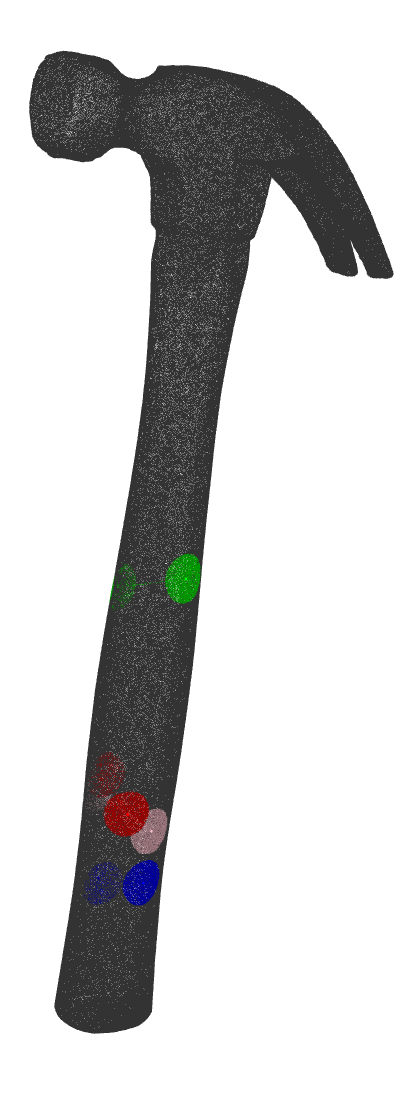}
    \end{minipage}

    \caption{Object meshes from LINEMOD and YCB-Video datasets used in experiments:
    (a) ape,
    (b) duck,
    (c) camera,
    (d) cat,
    (e) cracker box,
    (f) apple,
    (g) flat screwdriver,
    (h) mustard bottle,
    (i) power drill,
    (j) hammer.
    The enlarged sphere at the end of each line denotes the patch center, while the surrounding points are colored to indicate the spatial extent of the corresponding patch. For YCB object meshes, the points are smaller and more densely distributed, resulting in more compact patch centers and regions.}
    \label{fig:object_meshes}
\end{figure*}

\section{Experimental Results}

\subsection{Synthetic Test Set Experiments}
To evaluate the effectiveness of the proposed zero-shot grasp classifier, we conducted experiments on a synthetic test set comprising 500 graspable and 500 non-graspable surface patch pairs. The test data were randomly generated independently from the training and validation sets using the same procedural framework described in \ref{Graspable_patch}. 
The network was trained on the train partition of the synthetic dataset as described above. The system achieved a strong classification accuracy of $98\%$.

As a baseline, we implemented an analytical grasp feasibility classifier based on manually specified geometric criteria --- this geometric criteria was the same as that used to generate the synthetic dataset. When evaluated on the same synthetic test set, this
analytical method achieved an accuracy of $72.4\%$, which is far less than that of the learned model, despite the fact that the same geometric criteria were applied. This highlights a performance gap relative to the learned approach, due to augmentations applied after generating patch pairs, which increased the dataset's diversity and may have relaxed the criteria to some degree.

\subsection{Object Mesh Experiments}

To evaluate how the proposed approach transfers from isolated patch-pair classification that is not associated with any objects
to object-level grasp prediction, we conducted experiments on complete 3D object mesh models. These experiments assess whether a network trained solely on a purely synthetic dataset of procedurally generated local surface patch pairs can identify feasible grasps when applied exhaustively
across an object’s surface geometry.
By operating on full object meshes composed
of previously unseen geometries, this experiment directly tests the zero-shot generalization capability of the proposed method beyond the synthetic patch-pair training distribution.

\begin{figure}[t]
    \centering
    \includegraphics[width=0.5\textwidth]{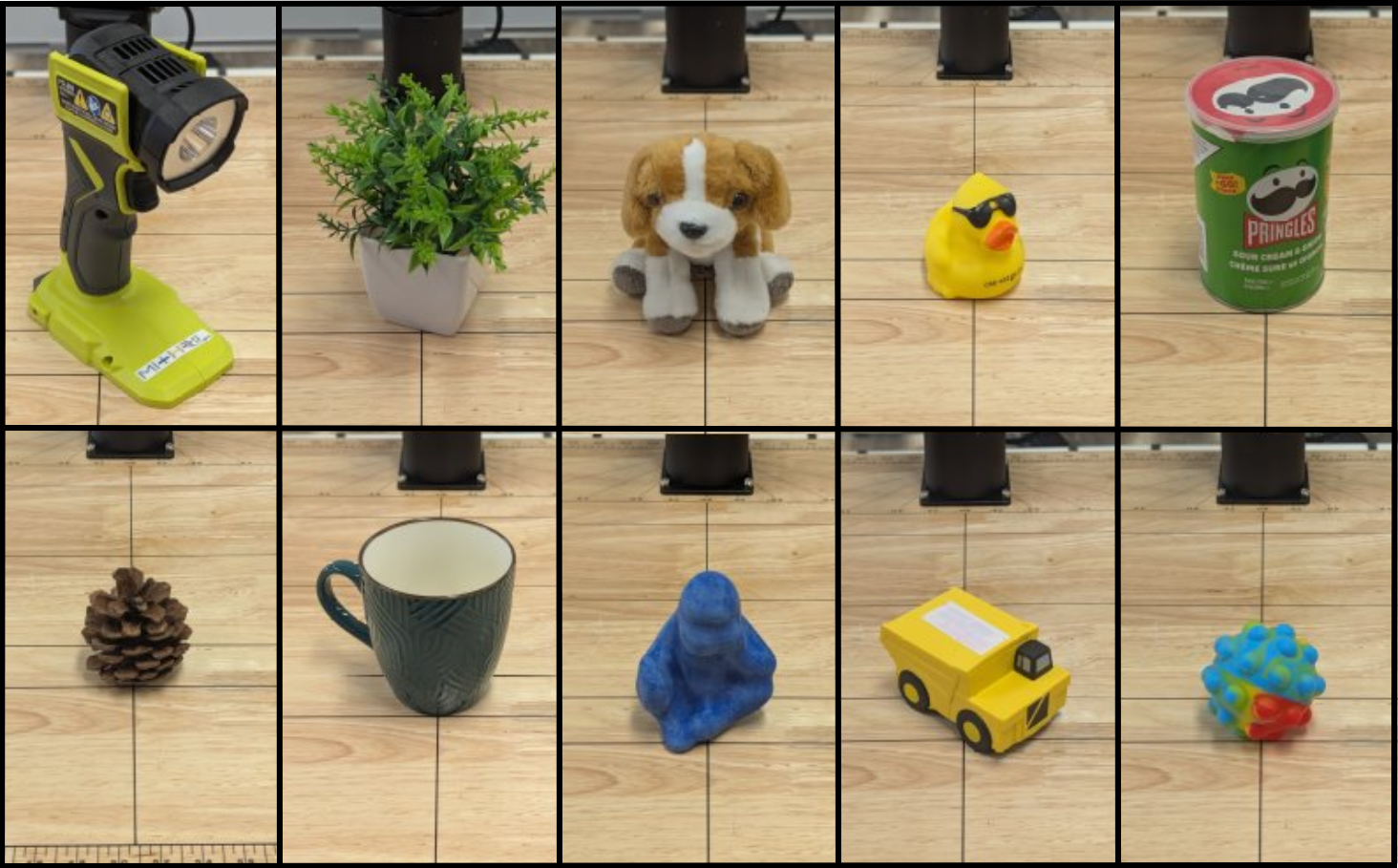}
    \caption{Objects used for real world testing}
    \label{fig:objs}
\end{figure}

The ten test objects include four from the LINEMOD dataset \cite{hinterstoisser2012model} (ape, duck, cat, camera)
and six from YCB-Video \cite{xiang2017posecnn} (apple, cracker box, mustard bottle, flat screwdriver, power drill, hammer), as shown in Figure~\ref{fig:objs}.
For each object mesh, 
200 patches are generated by randomly sampling 200 points from its mesh, and extracting a patch in the neighbourhood of each point using the same
$32\!\times\!32\!\times\!3$ XYZ representation employed during training. All patches are expressed in a common camera coordinate frame to preserve geometric consistency. Grasp candidates are generated by exhaustively pairing every extracted patch with every other patch from the same object, resulting in a set of 
$150\!\times\!149\!=\!22,350$
patch pairs that span a wide range of relative distances, orientations, and alignments that approximately cover the entire object surface. This exhaustive pairing strategy produces both geometrically plausible and implausible contact configurations, without relying on object identity, pose or symmetry.

Each patch pair is passed through the trained network, which outputs a scalar graspability score 
ranging from 0 (graspable) to 1 (non-graspable). Patch pairs with scores below a fixed threshold of 0.5 
are classified as graspable and retained as candidate grasps, while higher-scoring pairs are rejected as non-graspable. The resulting set of predicted grasps corresponds to pairs of surface locations on the object that satisfy the learned local geometric constraints of a parallel-jaw gripper. Only a small fraction of the patch-pair set is predicted as candidate grasps, with approximately 2,650 ($\sim\!1.2\%$) of all evaluated patch pairs classified as graspable across all objects. This strong sparsity reflects the selectivity of the learned geometric constraints and highlights the imbalance between graspable and non-graspable contact configurations at the object level.

To evaluate these results, we perform a qualitative reality check by visualizing the top-ranked 3 to 5 graspable patch pairs for each object,
as illustrated in Figure~\ref{fig:object_meshes}.
Non-maximum suppression~\cite{neubeck2006efficient} is applied to the top grasp candidates to remove redundant predictions arising from nearby patch pairs with similar spatial configurations and orientations. This step ensures that only geometrically distinct grasps are displayed, preventing clusters of nearly identical grasps from dominating the visualization and yielding a clearer representation of the model's object-level grasp predictions.



Qualitative inspection of the predicted grasps reveals several consistent trends.
Across most objects, the highest-ranked patch pairs correspond to antipodal or near-antipodal surface regions with locally planar or smoothly curved geometry, and with patch distances that are compatible with a parallel-jaw gripper.
On box-like objects such as the cracker box, predicted grasps predominantly lie on opposing planar faces, whereas on cylindrical or rounded objects such as the mustard bottle and apple, the predicted contacts tend to align along locally symmetric surface regions with antiparallel surface normals.
For elongated tools such as the flat screwdriver, hammer, and power drill, the model favours grasps along the handle region rather than highly concave functional ends with negligible patch distances, reflecting a bias toward locally consistent, low-curvature contact surfaces.
These observations demonstrate that the network learns generalizable zero-shot geometric grasps that are independent of object identity, rather than being object-specific.

To further assess the geometric validity of the predicted grasp candidates, we apply an algorithmic, analytical grasp-quality test to the subset of patch pairs classified as graspable (quality score $>95\%$) by the learned model.
For each predicted pair, we estimate local surface normals via PCA and evaluate whether the pair satisfies the same parallel-jaw grasping constraints used to generate the synthetic data, i.e., near-antipodal normal alignment, sufficient separation compatible with the gripper width, and approximately aligned contact points along the grasp axis. Patch pairs that meet all criteria are labelled as geometrically valid, while those that violate one or more constraints are rejected. This evaluation is used solely as an independent geometric consistency check and is not involved in training or threshold selection.
The results of this test are tabulated in Table~\ref{tab:classical_results}.


\begin{table}[t]
\centering
\caption{Algorithmic Geometric Evaluation of Network-Predicted Graspable Patch Pairs on YCB Objects}
\label{tab:classical_results}
\begin{tabular}{lccc}
\hline
\textbf{Object} & \textbf{Total Pairs} & \textbf{Good Pairs (\%)} & \textbf{Bad Pairs (\%)} \\
\hline
Cracker Box      & 125 & 67 (53.6\%) & 58 (46.4\%) \\
Mustard Bottle   & 70  & 15 (21.4\%) & 55 (78.6\%) \\
Flat Screwdriver & 27  & 3  (11.1\%) & 24 (88.9\%) \\
Apple            & 13  & 2  (15.4\%) & 11 (84.6\%) \\
Power Drill      & 88  & 7  (8.0\%)  & 81 (92.0\%) \\
Hammer           & 232 & 39 (16.8\%) & 193 (83.2\%) \\
\hline
\end{tabular}
\end{table}

The comparatively lower acceptance rates of the algorithmic geometric assessment across several objects can be attributed to fundamental limitations of rule-based grasp evaluation. The algorithmic method relies on a small set of hand-crafted geometric features and fixed thresholds on quantities such as patch extent, separation distance, and relative orientation. These hard constraints are inherently brittle to local curvature, discretization effects, and deviations from idealized geometric assumptions, which can limit their robustness across diverse object geometries. Thus, these results suggest that the analytic geometric criteria do not translate well to diverse real-world object geometries.

In contrast, the learned model evaluates graspability from the full point distributions of the patch pairs, allowing it to capture non-linear interactions between geometric features and tolerate moderate deviations from idealized antipodal contact. More importantly, extensive data augmentation was applied to both graspable and non-graspable patch pairs after their initial procedural generation, introducing increased variability in pose, orientation, and local geometry that is not explicitly expressed by the geometric constraints. Some of these augmentations also make the reliable computation of geometric quantities, such as patch distance and relative orientation, more diverse. 
While this augmentation improves the robustness and generalization of the learned model, it also highlights the sensitivity of fixed-threshold geometric tests across diverse object geometries.


\subsection{Real World Grasping Experiment}
\begin{figure}[htbp]
    \centering
    \begin{minipage}{0.48\textwidth}
        \centering
        \parbox{\linewidth}{\centering\includegraphics[width=\linewidth]{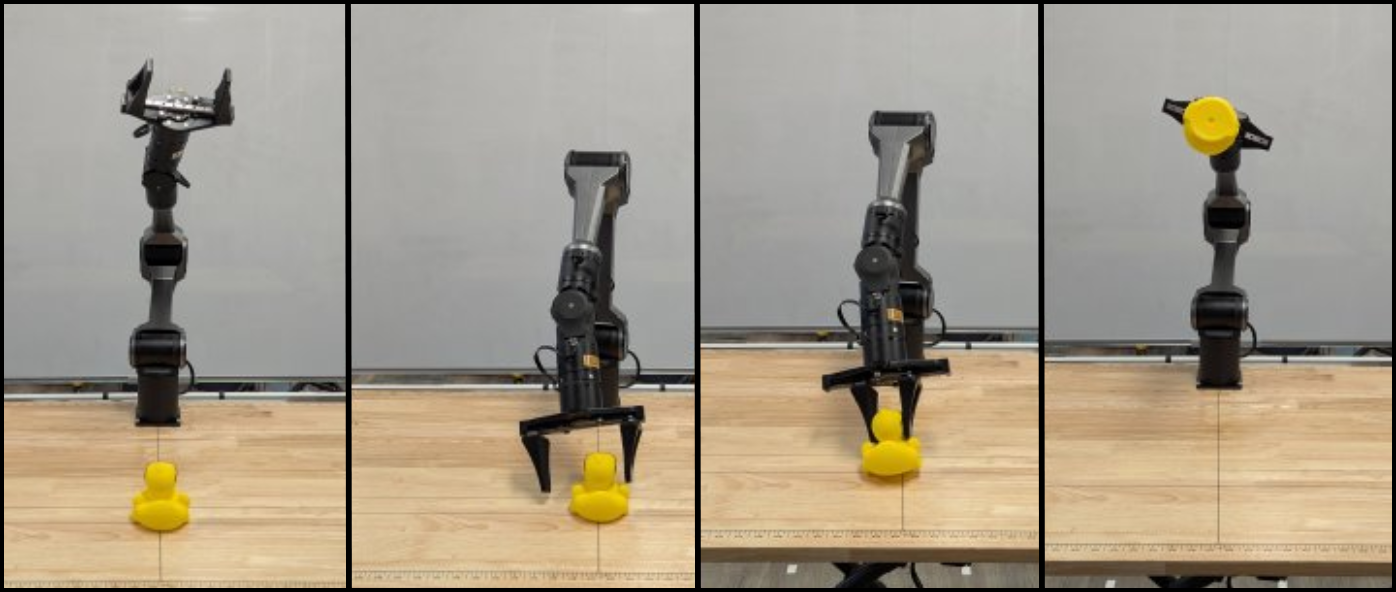}}

        \vspace{0.5em}

        \parbox{\linewidth}{\centering\includegraphics[width=\linewidth]{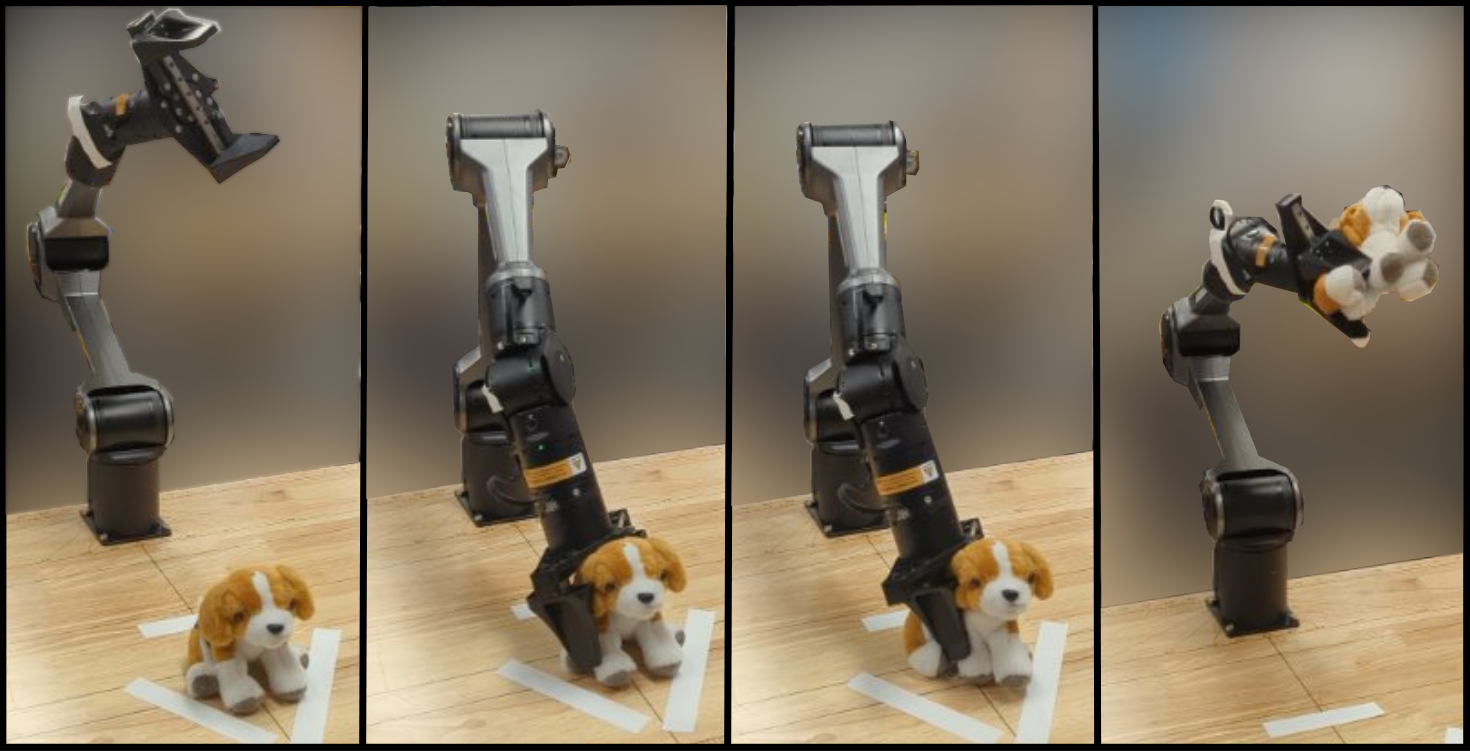}}

        \vspace{0.5em}

        \parbox{\linewidth}{\centering\includegraphics[width=\linewidth]{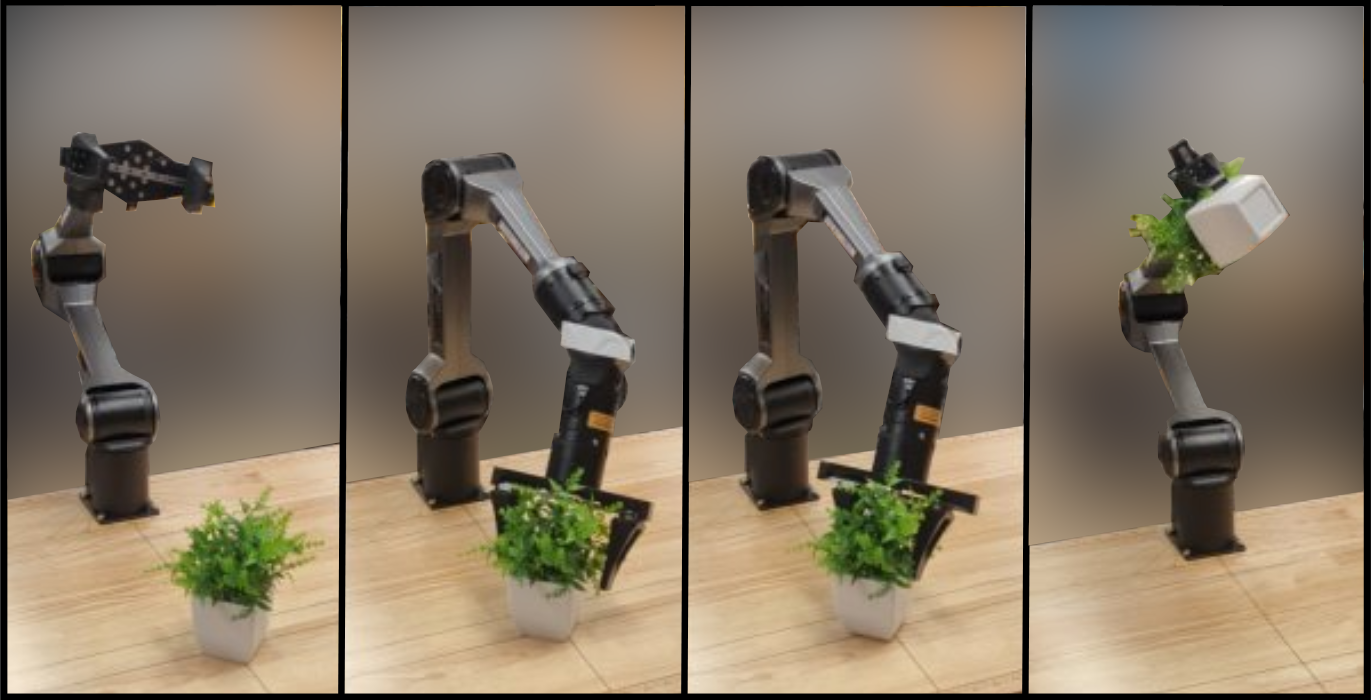}}
    \end{minipage}
    \caption{Grasping of objects using real-world data}
    \label{fig:grasps}
\end{figure}
\begin{figure*}[t]
    \centering
    \includegraphics[width=\textwidth]{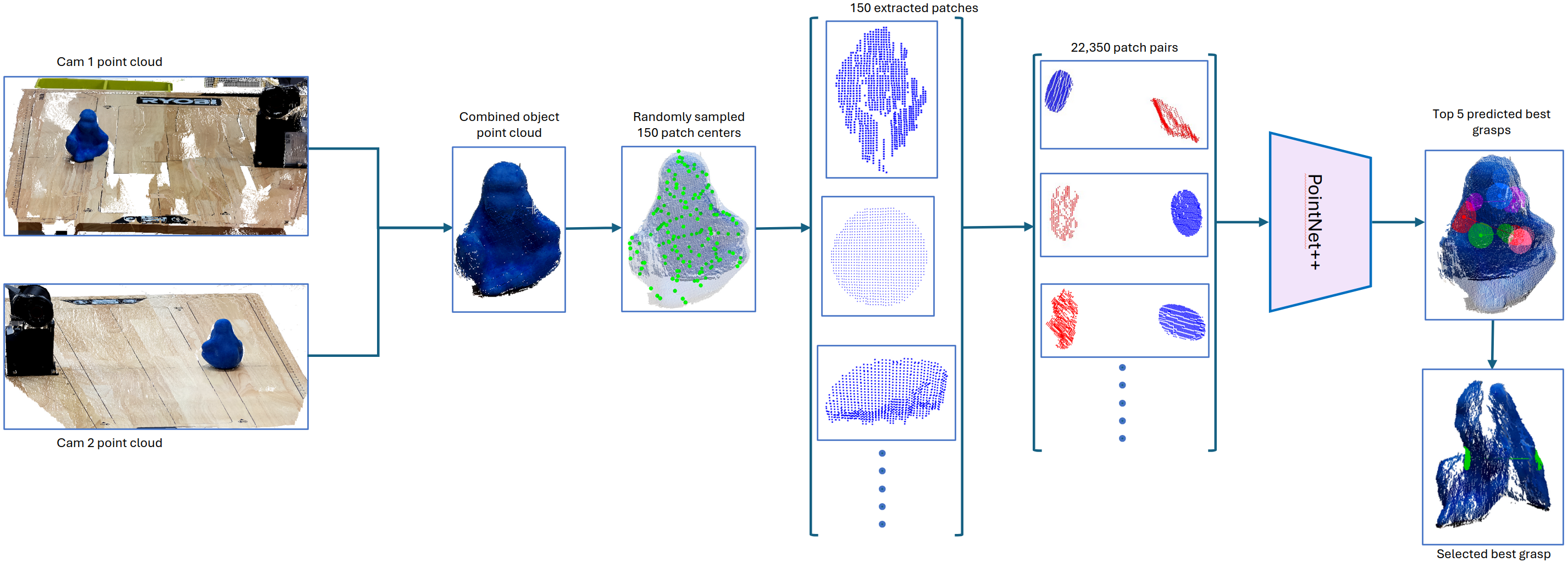}
    \caption{System pipeline.}
    \label{fig:system_pipeline}
\end{figure*}

To evaluate the model's effectiveness in a real-world setting, we conducted grasping experiments on 10 objects using an AgileX Piper 6-DOF robot arm \footnote{https://global.agilex.ai/products/piper}. Scene point clouds were acquired using a multi-view vision system consisting of two independent stereovision pairs. Each pair comprised two Raspberry Pi High Quality (HQ) cameras (utilizing the 12.3-megapixel Sony IMX477 sensor).
The stereovision pipeline was developed using the OpenCV library~\cite{bradski2000opencv}. For each camera pair, we performed a standard calibration procedure to determine intrinsic and extrinsic parameters, followed by stereo rectification to align the image planes. Disparity maps were computed using the StereoSGBM (Semi-Global Block Matching) algorithm~\cite{hirschmuller2008stereo} and then projected into 3D space. By registering the point clouds from both stereo pairs into a common coordinate frame, we generated a comprehensive 3D representation of the workspace. Finally, we sampled 150 grasp points (reduced the number of grasp point to improve computational efficiency) on each object’s surface and utilized the model to rank candidate grasp pairs. The highest-scoring pair was then passed to the grasp process described as follows. In case of ties, the pair whose midpoint was closest to the object center was selected as the grasp candidate. The overall perception-to-execution workflow is illustrated in~\ref{fig:system_pipeline}.

The grasp process takes a grasp candidate as two 3D points $p_1$ and $p_2$ in world coordinates, transforms them to the robot frame, and defines the grasp center as their midpoint and the gripper axis as the unit vector from $ p_1$ to $ p_2$. Approach directions are constrained to the plane perpendicular to this axis. The world $z$-axis is projected onto that plane to obtain an initial ``from above'' direction, then Rodrigues' formula rotates it about the gripper axis from $0$\textdegree{} to $180$\textdegree{} in $5$\textdegree{} increments; only directions with non-negative $z$ are considered. For each candidate, the desired end-effector pose (calculated by offsetting the position at the grasp center by 137 mm along the approach vector) is checked for reachability via an inverse-kinematics solver; the first reachable pose is selected, and the grasp is declared infeasible if none exists. Execution then moves to an initial configuration with the gripper open, advances to the chosen pose, closes the gripper, and then retracts to the initial configuration. 
Finally, the object is further rotated by $90$\textdegree{}
to ensure a stable grasp.

Of the ten objects evaluated, nine grasps were successful, and one grasp (the mug) was unreachable due to the arm's end-effector colliding with the object on approach. The grasps predicted by the model failed the kinematic test, suggesting the model struggles to reliably generate valid grasps on hollow objects.
Some examples of successful grasps are shown in Figure~\ref{fig:grasps}.

\section{Discussion and Conclusions}
Our paper demonstrates that reliable zero-shot grasp prediction can be achieved using local surface geometry alone. A model trained entirely on procedurally generated synthetic data was shown to generalize effectively to unseen object meshes and real-world grasping scenarios, without relying on object models or object-specific data like pose or identity. Across synthetic, object mesh, and real-world robot experiments, the system consistently identified geometrically plausible grasp contacts that align with classical parallel-jaw grasping principles.

The results highlight the strength of local, contact-centric representations for grasp feasibility assessment. By focusing on surface geometry at potential contact regions, the proposed approach avoids many sources of overfitting associated with object-aware or heavily supervised grasp pipelines, while remaining interpretable and transferable. This makes it well-suited for deployment in settings where object inventories are diverse or frequently changing such as warehouse automation, service robotics in homes or hospitals, and recycling or waste-sorting systems where object variability is high.

\textbf{Limitations and future work.}
The current approach evaluates geometric feasibility but does not explicitly model physical interaction effects, such as friction, compliance, or contact dynamics, which can lead to occasional failures under challenging conditions. While the representation is compatible with cluttered scenes, further work is needed to scale efficiently in dense environments and to reduce reliance on multi-view sensing. The method is currently limited to parallel-jaw grippers; extending the framework to multi-finger grasping will require defining appropriate geometric constraints and synthetic training strategies.

Overall, this study shows that local surface geometry provides a strong, transferable prior for zero-shot grasp-feasibility prediction, offering a lightweight and robust alternative to object-dependent grasping methods.


\bibliographystyle{IEEEtran}
\bibliography{ref}  

\end{document}